\documentclass{article} % For LaTeX2e
\usepackage{iclr2027_conference,times}

\usepackage{amsmath,amsfonts,bm}

\def\eqref#1{equation~\ref{#1}}
\def\1{\bm{1}}

\DeclareMathAlphabet{\mathsfit}{\encodingdefault}{\sfdefault}{m}{sl}
\SetMathAlphabet{\mathsfit}{bold}{\encodingdefault}{\sfdefault}{bx}{n}

\usepackage{hyperref}
\usepackage{url}
\usepackage{graphicx}
\usepackage{amsmath}
\usepackage{amssymb}
\usepackage{wrapfig}
\usepackage{booktabs}
\usepackage{pgfplots}
\pgfplotsset{compat=1.18}
\usepackage{multirow}
\usepackage{pifont}
\usepackage{caption}
\usepackage{subcaption}

\newcommand{\cmark}{\ding{51}}
\newcommand{\xmark}{\ding{55}}

\title{HiLo-WAM: Decoupling Long-Horizon Task Evolution and Local Dynamics in World Action Models}

\author{%
\parbox{\textwidth}{%
\raggedright
\normalfont
\vspace{2pt}
{\bfseries
\mbox{Xingyu Wang\textsuperscript{1,2}},
\mbox{Yanjiang Guo\textsuperscript{3}},
\mbox{Zhijing Ding\textsuperscript{1}},
\mbox{Jiaxin Fang\textsuperscript{4,1}},
\mbox{Zeyu Lai\textsuperscript{5,1}},
\mbox{Jiayu Dong\textsuperscript{1}},
\mbox{Weikang Wan\textsuperscript{1,5}},
\mbox{Jianyu Chen\textsuperscript{3}},
\mbox{Yuxiang Gao\textsuperscript{1,\ensuremath{\dagger}}},
\mbox{Mengdi Xu\textsuperscript{3,\ensuremath{\dagger}}}%
}\\[0.2em]
{\small
\mbox{\textsuperscript{1}CoCoMatrix}\quad
\textsuperscript{2}Institute of Automation, Chinese Academy of Sciences\quad
\mbox{\textsuperscript{3}Tsinghua University}\quad
\mbox{\textsuperscript{4}Fudan University}\quad
\textsuperscript{5}University of California, San Diego\quad
\mbox{\textsuperscript{\ensuremath{\dagger}}Corresponding author}
}\\[0.2em]
{\small
Project Page: \url{https://xywang227.github.io/HiLo-WAM/}\\
}%
}%
}

\iclrfinalcopy % Uncomment for camera-ready version, but NOT for submission.
\begin{document}

\maketitle
% \begingroup
% \renewcommand{\thefootnote}{\ensuremath{\dagger}}
% \footnotetext[1]{Corresponding author.}
% \endgroup
\begin{figure*}[h]
    \centering
    \vspace{-20pt}
    \includegraphics[width=\textwidth]{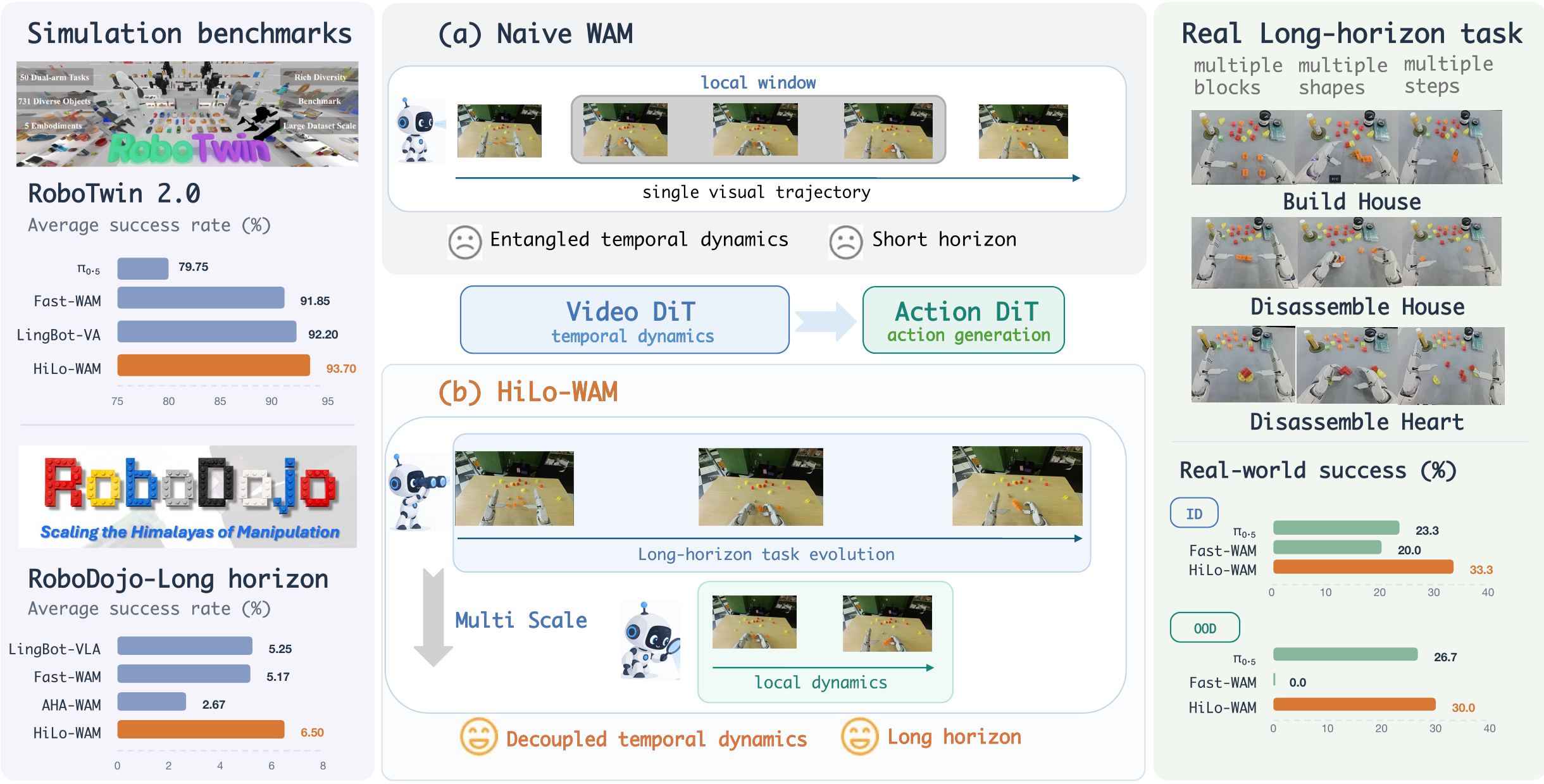}
    \vspace{-15pt}
    \caption{
    \textbf{Overview of HiLo-WAM.}
    Naive WAMs primarily model dense visual dynamics over a local interaction
    horizon, while HiLo-WAM decouples visual world modeling into a sparse long-horizon process
    for historical context and coarse future evolution and a dense local process
    for fine-grained interaction dynamics, and jointly models their interaction
    with actions.
    }
    \label{fig:motivation}
    \vspace{-5pt}
\end{figure*}
\begin{abstract}
World Action Models (WAMs) improve robotic manipulation by jointly modeling
visual dynamics and robot actions. However, most existing WAMs model dense
short-horizon visual dynamics, capturing local interactions but providing
limited long-range task context. Extending prediction horizons increases
uncertainty and computation while entangling task-level evolution with
fine-grained local dynamics, obscuring their dependencies and distinct
roles in action generation.
We propose \textbf{HiLo-WAM}, which decouples long-horizon task evolution
from local visual dynamics while jointly modeling their interaction
with robot actions. A sparse long-horizon visual stream captures historical
task context and future evolution, while a dense local dynamics visual stream models short-horizon robot--object interactions.
The two streams operate at different temporal scales within a unified
dual-DiT architecture and guide actions through structured cross-stream
attention. Progressive inference further reduces unnecessary computation through stream-wise partial denoising, joint denoising, and dynamic skipping.
Experiments demonstrate strong long-horizon manipulation performance:
HiLo-WAM achieves a $93.70\%$ average success rate on RoboTwin~2.0
and a $17.43$ progress score on RoboDojo long-horizon tasks, with robust
in-domain and OOD performance on the Fourier GR-3 robot.
Ablations confirm the complementary contributions of long-horizon
task evolution and local interaction dynamics, while temporal-scale
analyses highlight the importance of balancing temporal coverage
and resolution. Progressive inference further achieves a $2.93\times$ speedup while improving performance.
\end{abstract}

% \begin{figure*}[h]
%     \centering
%     \vspace{-17pt}
%     \includegraphics[width=\textwidth]{fig/overview.pdf}
%     \vspace{-15pt}
%     \caption{
%     \textbf{Overview of HiLo-WAM.}
%     Naive WAMs primarily model dense visual dynamics over a local interaction
%     horizon, while HiLo-WAM decouples visual world modeling into a sparse long-horizon process
%     for historical context and coarse future evolution and a dense local process
%     for fine-grained interaction dynamics, and jointly models their interaction
%     with actions.
%     }
%     \label{fig:motivation}
%     \vspace{-15pt}
% \end{figure*}
\section{Introduction}
\label{sec:intro}
Vision-Language-Action (VLA) models have shown strong generalization in robotic
manipulation through large-scale pretraining and diverse robot demonstrations
~\cite{pmlr-v305-black25a,black2026pi0visionlanguageactionflowmodel,
kim2025fine,intelligence2026pi07steerablegeneralistrobotic}. World Action
Models (WAMs) further introduce predictive visual dynamics into action
generation, either by conditioning actions on future visual states
~\cite{li2026causal,hu2024video,du2023learning} or by jointly modeling future
visual representations and actions~\cite{yuan2026fast,
li2025unifiedvideoactionmodel,ma2026dit4dit,ye2026world}. Most existing WAMs,
however, focus on dense visual dynamics over relatively short temporal windows,
which effectively capture fine-grained robot--object interactions but provide
limited information about how a task evolves over a longer horizon.
Long-horizon manipulation additionally requires understanding historical task
progress and future evolution beyond the current interaction window~\cite{cai2026aha,hu2026bagelvla,torne2026memmultiscaleembodiedmemory,yang2026memorywam}.

To incorporate such long-range task information, recent WAMs extend the
temporal scope of visual modeling through longer observation histories~\cite{li2026causal,ye2026world},
persistent memory~\cite{yang2026memorywam}, extended
future prediction~\cite{xu2026forcingcausalworldmodeling}, or long-horizon visual planning~\cite{cai2026aha}. These
approaches provide richer temporal context by exposing the policy to task
progress and future dynamics over a broader time span.
% Vision-Language-Action (VLA) models have achieved strong generalization in
% robotic manipulation through large-scale pretraining and diverse robot
% demonstrations~\cite{pmlr-v305-black25a,
% black2026pi0visionlanguageactionflowmodel,kim2025fine,
% intelligence2026pi07steerablegeneralistrobotic}.
% World Action Models (WAMs) extend this paradigm by jointly modeling visual
% world dynamics and robot actions, injecting predictive world information into
% policy learning. They either predict future visual states to condition action
% generation~\cite{li2026causal,hu2024video,du2023learning}, or jointly denoise
% future visual representations and actions~\cite{yuan2026fast,
% li2025unifiedvideoactionmodel,ma2026dit4dit,ye2026world}.
% This explicit modeling of world evolution provides a promising foundation for
% generalizable and long-horizon manipulation.

% Most existing WAMs, however, predict dense visual dynamics over relatively
% short temporal windows. Such predictions provide rich supervision for
% fine-grained robot--object interactions, making them well suited for modeling
% \emph{local interaction dynamics}. Long-horizon manipulation requires an
% additional capability: understanding how the task evolves beyond the immediate
% interaction window, including historical task progress and longer-term future
% development. Recent WAMs therefore introduce longer historical context,
% persistent memory~\cite{ye2026world,li2026causal,yang2026memorywam}, extended
% future prediction, or long-horizon visual planning~\cite{cai2026aha} to
% incorporate longer-range task information.

Extending the temporal horizon of visual world modeling introduces a
fundamental challenge: a single visual trajectory must capture both slowly
evolving task-level progression and fine-grained local interaction dynamics.
This not only increases prediction uncertainty and redundant computation~\cite{agibotresearchteam2026geact20pretrainingscaling,cai2026aha,zhou2025act2goalworldmodelgeneral}, but,
more importantly, obscures the dependencies between different
temporal dynamics and their distinct roles in action generation. This raises a
key question: \emph{how can a World Action Model explicitly model different
temporal dynamics and their dependencies on action generation?}

% \begin{figure*}[t]
%     \centering
%     \includegraphics[width=\textwidth]{fig/overview.pdf}
%     \vspace{-15pt}
%     \caption{
%     \textbf{Overview of HiLo-WAM.}
%     Standard WAMs primarily model dense visual dynamics over a local interaction
%     horizon, while long-horizon manipulation additionally requires understanding
%     task-level evolution over extended time.
%     HiLo-WAM decouples visual world modeling into a sparse long-horizon process
%     for historical context and coarse future evolution and a dense local process
%     for fine-grained interaction dynamics, and jointly models their interaction
%     with actions.
%     }
%     \label{fig:motivation}
%     \vspace{-15pt}
% \end{figure*}

To address this problem, we propose \textbf{HiLo-WAM}, which decouples the two
temporal roles of visual world modeling into complementary long-horizon and
local dynamics streams while jointly learning their interaction with robot
actions. As shown in Fig.~\ref{fig:motivation}, We adopt a \emph{multi-scale temporal modeling} strategy to represent
the two streams at different temporal resolutions. Specifically, the
\emph{long-horizon visual stream} sparsely models historical task context and
coarse future task evolution, capturing how the task has progressed and where
it is heading, while the \emph{local visual dynamics stream} densely models
short-horizon robot--object interactions required for precise control. Both
streams are processed by a shared Video DiT and interact with an Action DiT
through structured cross-stream attention, allowing long-horizon task
information and local dynamics to jointly support action generation. This
temporal decomposition further enables progressive stream-wise inference,
where visual streams terminate and are cached at different stages to reduce
unnecessary world-model computation.

We evaluate HiLo-WAM in simulation and real-robot tasks. HiLo-WAM achieves state-of-the-art performance on RoboTwin 2.0~\cite{chen2025robotwin} and strong performance on challenging long-horizon manipulation tasks~\cite{chen2026robodojounifiedsimandrealbenchmark}. Controlled ablations and attention analyses show that long-horizon stream with historical context, future evolution, and dense local visual dynamics provide complementary information for action generation. Temporal-scale analyses further demonstrate that long-horizon world modeling benefits from an appropriate temporal scale that balances temporal coverage and resolution. Progressive inference further achieves a $2.93\times$ speedup while improving performance.

Our contributions are summarized as follows:
\begin{itemize}
% \vspace{-7.7pt}
    \item We propose \textbf{HiLo-WAM}, a World Action Model that decouples
    visual world modeling into complementary long-horizon and local dynamics
    processes, and jointly exploits both for robot action generation.

    \item We introduce a multi-scale temporal modeling strategy with structured
    cross-stream interaction and progressive stream-wise inference, enabling
    effective coordination across temporal scales while reducing unnecessary
    world-model computation.

    \item Extensive simulation and real-world experiments demonstrate strong
long-horizon manipulation performance, with ablations and further analyses validate the benefit of complementary temporal information, the importance of balancing temporal coverage and resolution, and the efficiency of progressive inference.
\end{itemize}
\section{Related Works}
\label{sec:related}

\textbf{Video Models and World Action Models.}
Recent advances in video generation have enabled visual prediction to serve as
a powerful dynamics prior for robot learning~\cite{black2026pi0visionlanguageactionflowmodel,Wang_2026_CVPR,hu2024video}. Prior works use video models to
generate future observations for planning and control~\cite{
hu2024video,du2023learning,li2026causal}, or jointly model future visual
states and robot actions within a unified World Action Model
(WAM)~\cite{yuan2026fast,li2025unifiedvideoactionmodel,
ma2026dit4dit,ye2026world}. While early WAMs mainly focus on short-horizon
interaction dynamics, recent approaches incorporate longer context,
persistent memory, or long-range future prediction and visual
planning~\cite{ye2026world,yang2026memorywam,cai2026aha,zhou2025act2goalworldmodelgeneral,agibotresearchteam2026geact20pretrainingscaling}
to support more complex long-horizon manipulation. 
HiLo-WAM instead decouples long-horizon manipulation into two complementary visual streams: a long-horizon stream for long-horizon task evolution and a local stream for fine-grained interaction dynamics.

\textbf{Hierarchical Policies for Long-Horizon Manipulation.}
A parallel line of work addresses long-horizon manipulation through
hierarchical or temporally abstract policies. These methods typically use
high-level language plans, subgoals, visual goals, or intermediate states to
capture task-level structure, and condition a downstream policy on these
representations to generate fine-grained actions~\cite{
shi2025hi,li2025hamster,long2026scaling,pmlr-v305-black25a,
yang2026hivla,intelligence2026pi07steerablegeneralistrobotic,
11093669,hu2026bagelvla}.
HiLo-WAM shares the motivation of combining long-range reasoning with precise
local control, but approaches it through visual world modeling rather than a
separate planner--policy hierarchy. It decouples the visual dynamics of a
unified WAM into long-horizon and local streams and jointly learns their
interaction with robot actions.

\section{Method}
\label{sec:method}

\begin{figure*}[t]
    \centering
    \includegraphics[width=\textwidth]{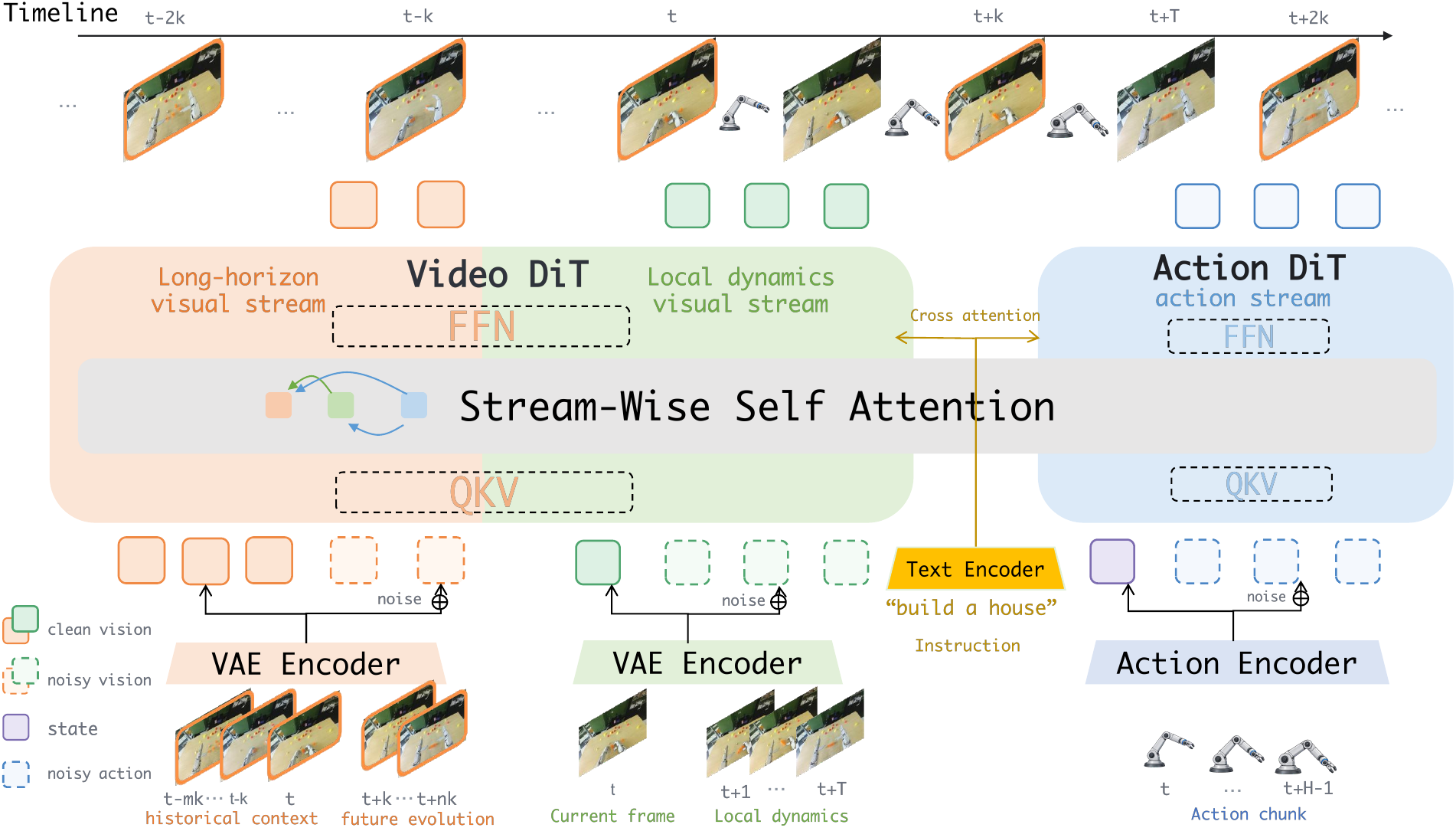}
    \vspace{-15pt}
    \caption{
    \textbf{Architecture of HiLo-WAM.}
    HiLo-WAM builds on a dual-DiT World Action Model consisting of a shared
    Video DiT and an Action DiT.
    The visual world is modeled at two complementary temporal scales:
    a sparse long-horizon visual stream captures historical context and coarse
    future evolution, while a dense local visual dynamics stream captures
    fine-grained local interactions.
    Both visual streams are jointly modeled with robot actions.
    During inference, the visual streams progressively terminate and are cached,
    while the action stream continues denoising to completion.
    }
    \label{fig:framework}
    \vspace{-15pt}
\end{figure*}

\paragraph{Overview.}
We introduce \textbf{HiLo-WAM}, a World Action Model that decouples
visual world modeling into two complementary visual streams:
a \emph{long-horizon visual stream} which sparsely models historical task context
and coarse future evolution for task-level evolution and a dense
\emph{local visual dynamics stream} for fine-grained interaction dynamics.
HiLo-WAM adopts a \emph{multi-scale temporal modeling} strategy, representing
the two visual streams at different temporal scales. As illustrated in
Fig.~\ref{fig:framework}, the model follows a dual-DiT architecture consisting
of a Video DiT for visual world modeling and an Action DiT for action stream
generation. Both visual streams are jointly modeled by
the shared Video DiT and interact with action stream through structured
cross-stream attention, allowing long-horizon task evolution and local dynamics
to jointly guide action prediction. During inference, all streams are
progressively denoised: the long-horizon visual stream terminates and is cached
first, followed by the local visual stream, while the action stream is refined
to completion.

% ============================================================
\subsection{Problem Formulation}
\label{sec:wam}

At control step $t$, the robot observes an image $o_t$, a proprioceptive
state $s_t$, and a language instruction $l$, and predicts an action chunk
$A_t=(a_t,\ldots,a_{t+H-1})$. Conventional World Action Models (WAMs)
augment action prediction with predictive visual dynamics by jointly modeling
$p_\theta(V_t,A_t \mid o_t,s_t,l)$, where
$V_t=(o_{t+1},\ldots,o_{t+T})$ denotes a future visual trajectory.
For long-horizon manipulation, this formulation is extended by incorporating a
longer visual history $O_t=(o_{t-M},\ldots,o_t)$ and a longer future trajectory
$V_t=(o_{t+1},\ldots,o_{t+T})$ to provide richer historical context and capture task evolution over a longer temporal span, yielding
$p_\theta(V_t,A_t \mid O_t,s_t,l)$, where $M$ and $T$ control the historical
and predictive temporal horizons, respectively.

\subsection{Multi-scale Temporal Modeling}
\label{sec:mstm}
HiLo-WAM decouples visual world modeling into two complementary streams using
a multi-scale temporal representation. 
Just like the standard WAM, the local visual dynamics stream
$V_t^{\mathrm{L}}=\{o_{t+j}\}_{j=0}^{T}$
learns to densely capture fine-grained local interactions.
The long-horizon visual stream
$V_t^{\mathrm{H}}=\{o_{t+ik}\}_{i=-m}^{n}$
downsamples observations with temporal stride $k$ relative to local stream, providing broader but sparser
temporal coverage for historical task context and coarse future evolution.
HiLo-WAM jointly models
\begin{equation}
    p_\theta(V_{t,i>0}^{\mathrm{H}}, V_{t,j>0}^{\mathrm{L}}, A_t
    \mid V_{t,i\leq0}^{\mathrm{H}}, V_{t,j=0}^{\mathrm{L}}, s_t,l),
    \label{eq:hilo_formulation}
\end{equation}
allowing sparse historical task context and current state to predict coarse future evolution, local dynamics and action chunk.

% ============================================================
\subsection{Model Architecture}
\label{sec:architecture}

\paragraph{Architecture.}
HiLo-WAM follows the Mixture-of-Transformers (MoT) formulation of recent
World Action Models~\cite{cai2026aha,li2026causal,yuan2026fast}. The model consists of a Video DiT for visual world
modeling and an Action DiT for robot action generation. The two branches
retain modality-specific parameters and exchange information through
layer-wise joint attention. We initialize the Video DiT and its video VAE
from Wan2.2-5B. A pretrained T5 encoder provides language conditioning to
both branches, while a action encoder maps the proprioceptive state $s_t$ and action chunk $A_t$
to a state token $q_t$ and action tokens $Z_t^{\mathrm{A}}$ for the Action DiT.

Given the long-horizon and local visual streams defined in
Sec.~\ref{sec:mstm}, we encode them independently using the same pretrained
video VAE $\mathcal{E}_v$ to preserve their individual temporal structure:
\begin{equation}
    Z_t^{\mathrm{H}}
    =\mathcal{E}_v(V_t^{\mathrm{H}}),
    \qquad
    Z_t^{\mathrm{L}}
    =\mathcal{E}_v(V_t^{\mathrm{L}}),
    \label{eq:video_encoding}
\end{equation}
where $Z_t^{\mathrm{H}}$ and $Z_t^{\mathrm{L}}$ denote the long-horizon and
local visual latent sequences, respectively. 
Finally, We organize the input tokens into four groups:
\begin{equation}
    \mathcal{Z}_t
    =
    \left[
        Z_t^{\mathrm{H}},
        Z_t^{\mathrm{L}},
        q_t,
        Z_t^{\mathrm{A}}
    \right],
    \label{eq:token_groups}
\end{equation}
During training, latent tokens associated with history context and current
observation remain clean conditions, while future visual and action tokens
are noised and predicted. Information exchange among the four groups is
controlled by the structured attention mechanism introduced next.

\paragraph{Structured cross-stream attention.}
% \vspace{-10pt}
% \begin{wrapfigure}[16]{r}{0.4\textwidth}
%     \centering
%     \vspace{-25pt}
%     \includegraphics[width=\linewidth]{fig/attention.pdf}
%     \vspace{-17pt}
%     \caption{
%     \textbf{Cross-stream attention.}
%     }
%     \label{fig:attention}
% \end{wrapfigure}
After the video VAE performs spatiotemporal compression, 
we partition the latent video of the long-horizon stream into three parts:
\begin{equation}
Z_t^{\mathrm H}
=
\left[
Z_t^{\mathrm{H,ctx}},
Z_t^{\mathrm{H,near}},
Z_t^{\mathrm{H,far}}
\right],
\end{equation}
where $Z_t^{\mathrm{H,ctx}}$ represents the latent historical context,
$Z_t^{\mathrm{H,near}}$ corresponds to the nearest latent future frame, and
$Z_t^{\mathrm{H,far}}$ represents more distant latent future frames.

\begin{wrapfigure}[16]{r}{0.4\textwidth}
    \centering
    \vspace{-20pt}
    \includegraphics[width=\linewidth]{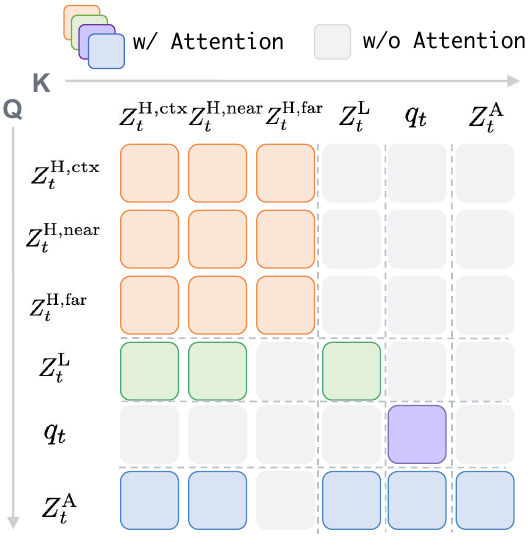}
    \vspace{-17pt}
    \caption{
    \textbf{Cross-stream attention.}
    }
    \label{fig:attention}
\end{wrapfigure}

As shown in Fig.~\ref{fig:attention}, HiLo-WAM employs an asymmetric
cross-stream attention mask. Long-horizon latent tokens attend bidirectionally
within the long-horizon stream. Local visual latents can attend to
$Z_t^{\mathrm{H,ctx}}$, $Z_t^{\mathrm{H,near}}$, and the local stream itself,
allowing task-level context and near-term coarse evolution to guide
fine-grained dynamics modeling. Action tokens additionally attend to the
proprioceptive state token and action tokens themselves, while receiving
information from both visual streams.

The more distant future latents $Z_t^{\mathrm{H,far}}$ are not directly
accessible to the local visual or action streams, preventing local prediction
and control from relying directly on increasingly uncertain long-range visual
states. Visual tokens do not attend to action tokens, while the state token
$q_t$ remains a clean condition.

% The three token groups interact through the structured attention mask shown in
% Fig.~\ref{fig:attention}. Within each group, bidirectional attention
% supports coherent temporal modeling of the corresponding modality.
% The local visual stream and action tokens can attend to the observed long-horizon
% context and the nearest coarse future prediction, allowing task-level
% information to guide local dynamics modeling and action generation.
% More distant future predictions remain internal to the long-horizon stream and participate only in bidirectional long-horizon planning. They are not directly exposed to the local visual or action streams, avoiding excessive dependence of local prediction and control on uncertain long-range visual states.

% Cross-stream information flow is therefore organized from long-horizon task
% evolution to local interaction dynamics and ultimately to action generation.
% The proprioceptive state token remains a clean condition and attends only to
% itself, while action tokens can attend to the state. Visual tokens do not
% attend to action tokens. This asymmetric attention structure preserves the
% complementary roles of the two visual streams and also allows upstream
% visual representations to be frozen and cached during progressive inference.

% ============================================================
\subsection{Joint Flow-Matching Objective}
\label{sec:training}

HiLo-WAM jointly optimizes long-horizon and local dynamics visual prediction, and action generation with flow matching. For a clean
target $y$ from any prediction stream, we sample Gaussian noise
$\epsilon\sim\mathcal{N}(0,I)$ and a flow time $\tau\in[0,1]$, and construct
the interpolated sample
$y_{\tau}
    =
    (1-\tau)y+\tau\epsilon .$
% \begin{equation}
%     y_{\tau}
%     =
%     (1-\tau)y+\tau\epsilon .
%     \label{eq:flow_path}
% \end{equation}
The model predicts the corresponding velocity field
$u=\epsilon-y$ and is trained with
\begin{equation}
    \mathcal{L}_{\mathrm{FM}}(y)
    =
    \mathbb{E}_{y,\epsilon,\tau}
    \left[
        \left\|
        f_{\theta}(y_{\tau},\tau,\mathcal{C})-(\epsilon-y)
        \right\|_2^2
    \right],
    \label{eq:fm_objective}
\end{equation}
where $\mathcal{C}$ denotes the clean visual context, proprioceptive state,
language instruction, and the accessible cross-stream context determined by
the structured attention mask.

We instantiate this objective for the future portion of the long-horizon
visual stream $\mathcal{L}_{\mathrm{H}}
        = \mathcal{L}_{\mathrm{FM}}
        \bigl(Z_{t,i>0}^{\mathrm{H}}\bigr)$,
the future portion of the local visual stream $\mathcal{L}_{\mathrm{L}}
        = \mathcal{L}_{\mathrm{FM}}
        \bigl(Z_{t,j>0}^{\mathrm{L}}\bigr)$, and the action
tokens $\mathcal{L}_{\mathrm{A}}
        = \mathcal{L}_{\mathrm{FM}}
        \bigl(Z_t^{\mathrm{A}}\bigr)$.
% \begin{equation}
% \begin{aligned}
%     \mathcal{L}_{\mathrm{H}}
%         &= \mathcal{L}_{\mathrm{FM}}
%         \bigl(Z_{t,i>0}^{\mathrm{H}}\bigr),\qquad
%     \mathcal{L}_{\mathrm{L}}
%         &= \mathcal{L}_{\mathrm{FM}}
%         \bigl(Z_{t,j>0}^{\mathrm{L}}\bigr), \qquad
%     \mathcal{L}_{\mathrm{A}}
%         &= \mathcal{L}_{\mathrm{FM}}
%         \bigl(Z_t^{\mathrm{A}}\bigr).
% \end{aligned}
% \label{eq:stream_losses}
% \end{equation}
The overall training objective is
\begin{equation}
    \mathcal{L}
    =
    \lambda_{\mathrm{H}}\mathcal{L}_{\mathrm{H}}
    +
    \lambda_{\mathrm{L}}\mathcal{L}_{\mathrm{L}}
    +
    \lambda_{\mathrm{A}}\mathcal{L}_{\mathrm{A}},
    \label{eq:joint_loss}
\end{equation}
where $\lambda_{\mathrm{H}}$, $\lambda_{\mathrm{L}}$, and
$\lambda_{\mathrm{A}}$ balance learning across different streams, respectively. We independently sample the flow
times for the three streams, allowing them to be trained under different noise
levels and supporting the progressive denoising strategy used during inference.

% ============================================================
\subsection{Efficient Progressive Inference}
\label{sec:inference}

Introducing multi-scale visual streams adds denoising computation and
increases inference latency. To reduce this overhead while maintaining
manipulation performance, we combine three complementary strategies:
partial denoising, joint denoising, and dynamic step skipping.

\noindent\textbf{Partial denoising.}
Useful dynamics information can emerge before visual predictions are
fully denoised~\cite{hu2024video,li2026causal}.
We assign denoising budgets $n_{\mathrm H}$, $n_{\mathrm L}$, and
$n_{\mathrm A}$ to the long-horizon, local dynamics, and action streams,
respectively, with $n_{\mathrm H}\leq n_{\mathrm L}\leq n_{\mathrm A}$.
Each visual stream terminates upon reaching its budget and caches its
features, while the action stream completes its full denoising schedule.
The structured asymmetric attention naturally supports this scheme,
allowing downstream streams to reuse cached upstream representations.

\noindent\textbf{Joint denoising.}
To better align representations across denoising stages and further
accelerate inference, we start denoising all streams simultaneously.
During early denoising, action and visual representations interact
at the same denoising timestep through structured cross-stream attention.
During later denoising, the action stream uses cached visual
representations that capture informative dynamics to refine its
predictions and generate precise actions.

\noindent\textbf{Dynamic skipping.}
Following DreamZero~\cite{ye2026world}, we exploit temporal redundancy across
denoising steps. When consecutive predictions $v_t$ and $v_{t-1}$ at denoising step $t$ have cosine similarity above a predefined threshold $\gamma$, We therefore approximate
the next $k$ velocity predictions using $v_t$, i.e.,
$\hat{v}_{t+j}\approx v_t$ for $j=1,\ldots,k$,
skipping the corresponding DiT evaluations to accelerate inference.

\section{Experiments}
\label{sec:experiments}

We conduct experiments in simulation and on a real-world Fourier GR-3 robot
to investigate three questions:
\textbf{(1)}
    Does HiLo-WAM improve long-horizon manipulation performance compared with existing VLA and World Action Models?
\textbf{(2)}
    How does visual dynamics at different temporal scales contribute to action generation?
\textbf{(3)}
    How does the temporal scale affect manipulation performance?
\textbf{(4)}
    Can progressive inference improve inference efficiency while maintaining performance?

% We evaluate HieWAM in simulation and on a physical robot to answer three
% questions. \textbf{Q1: Overall capability.} Does hierarchical world--action
% modeling improve long-horizon manipulation over strong VLA and WAM baselines?
% \textbf{Q2: Visual-stream contribution.} What does each visual stream contribute
% to long-horizon control, and are the long-horizon and local-dynamics streams
% complementary? \textbf{Q3: Interaction mechanism.} How do the long-horizon
% visual stream, local visual dynamics stream, and action stream interact during
% prediction? We first study these questions on RoboTwin~2.0 and then evaluate
% real-world deployment, task progress, and inference efficiency on a Fourier
% GR-3 robot.

\subsection{Experimental Setup}
\label{sec:exp_setup}

\paragraph{RoboTwin 2.0.}
RoboTwin~2.0 is a large-scale bimanual manipulation benchmark containing 50
dual-arm tasks under both clean and domain-randomized evaluation
~\citep{chen2025robotwin}. We follow the multi-task protocol of Fast-WAM
~\citep{yuan2026fast}: the training set contains 50 clean demonstrations and
500 randomized demonstrations per task, totaling 2,500 clean and 25,000
randomized trajectories. We evaluate each task over 50 trials in both settings
and report success rate averaged across all 50 tasks. The randomized setting
varies scene clutter, illumination, background texture, tabletop configuration,
and other visual or physical factors.

\paragraph{RoboDojo Benchmark.}
RoboDojo~\cite{chen2026robodojounifiedsimandrealbenchmark} is a general robotic manipulation benchmark consisting of 42 tasks spanning five evaluation dimensions. We follow the official training and evaluation setup and train our model on 3,500 demonstration episodes. To assess long-horizon manipulation performance, we evaluate on eight long-horizon tasks in RoboDojo, with 50 trials per task. We report both task progress scores and success rates
\footnote{Baseline results are taken from the official RoboDojo simulation leaderboard. To ensure a fair comparison, We exclude certain models that are not open-source or have
undergone large-scale embodied pretraining.}.

\paragraph{Real-world platform and tasks.}
We deploy policies on a Fourier GR-3 humanoid robot. We consider three
long-horizon bimanual tasks: \emph{Build House}, \emph{Disassemble House},
\emph{Disassemble Heart}. The assembly tasks require
the policy to maintain a target structure over multiple placements, whereas the
disassembly tasks require progress tracking while reversing the structure into
its constituent blocks. We collect 50 hours of teleoperated demonstrations in
total and train each model for five epochs on the same task-specific data.

% We compare HiLo-WAM with $\pi_{0.5}$~\citep{pmlr-v305-black25a} and
% Fast-WAM~\citep{yuan2026fast}. Each method is evaluated over 10 trials per
% task. A rollout is successful if the final task state is completed within the
% fixed episode horizon. To capture partial progress, we additionally decompose
% each task into ten ordered milestones and assign a score in
% $\{0-10\}$ according to the number of completed milestones. We report mean
% success rate and mean progress score over all trials. The exact task definitions
% and milestone criteria are provided in the appendix.

\paragraph{Implementation details.}
HiLo-WAM is initialized from the same pretrained video backbone
~\citep{wan2025wan} used by the closest WAM baseline
~\citep{yuan2026fast}. We use logit-normal flow-time
sampling, AdamW with learning rate $1\times10^{-4}$ and weight decay $0.01$, a
cosine learning-rate schedule, mixed-precision training, and gradient clipping
at $1.0$~\citep{yuan2026fast}. We use 10 action denoising steps with
classifier-free guidance scale $1.0$. The temporal configuration
$(k,m,n,T,H)$, stream-specific stopping steps $(n_H,n_L,n_A)$, loss weights are reported in the appendix~\ref{app:training}.

\subsection{Simulation Performance Analysis (Q1)}
\label{sec:robotwin_results}

\paragraph{Quantitative Analysis.}
% \begin{wraptable}{r}{0.57\textwidth}
% \vspace{-45pt}
% \centering
% \small
% \setlength{\tabcolsep}{5pt}
% \renewcommand{\arraystretch}{1.0}

% \begin{tabular}{lcc}
% \toprule
% \textbf{Model} & \textbf{Score} & \textbf{Success (\%)} \\
% \midrule

% \multicolumn{3}{l}{\textit{VLA-based methods}} \\
% OpenVLA-OFT~\citep{kim2025fine}  & 0.70  & 0.00 \\
% RDT~\citep{liu2025rdt}          & 1.13  & 0.00 \\
% LDA-1B~\citep{lyu2026lda}       & 1.98  & 0.08 \\
% H-RDT~\citep{bi2025hrdthumanmanipulationenhanced}        & 2.23  & 0.17 \\
% Pi-0~\citep{black2026pi0visionlanguageactionflowmodel}         & 6.19  & 2.00 \\
% GROOT-N1.7~\citep{bjorck2025gr00t}   & 8.30  & 3.58 \\
% LingBot-VLA~\citep{wu2026pragmatic}  & 10.89 & 5.25 \\
% StarVLA~\citep{community2026starvlalegolikecodebasevisionlanguageaction}      & 14.15 & \textbf{6.50} \\
% StarVLA-OFT  & \underline{14.28} & \underline{6.17} \\

% \midrule
% \multicolumn{3}{l}{\textit{World-model-based methods}} \\
% AHA-WAM~\citep{cai2026aha}      & 8.61 & 2.67 \\
% Fast-WAM~\citep{yuan2026fast}     & 9.14 & 5.17 \\

% \midrule
% \textbf{HiLo-WAM}
% & \textbf{17.43}
% & \textbf{6.50} \\

% \bottomrule
% \end{tabular}
% \vspace{-7pt}
% \caption{
% \textbf{RoboDojo long-horizon evaluation.}
% Score and average success rate (\%) across 8 long-horizon tasks.
% Best and second-best values among the listed
% methods are shown in bold and underlined, respectively.}
% \label{tab:robodojo_main}
% \vspace{-25pt}
% \end{wraptable}
Tables~\ref{tab:main_results}(a) and~\ref{tab:main_results}(b) summarize the results on RoboTwin~2.0 and long-horizon tasks in RoboDojo. Overall, HiLo-WAM achieves performance comparable to state-of-the-art methods on both benchmarks without using embodied pretraining.

On RoboTwin~2.0, HiLo-WAM achieves the strongest result among the compared methods, with success
rates of $93.87\%$ and $93.52\%$ in the clean and randomized settings,
respectively. Its average success rate of $93.70\%$ improves over Fast-WAM by
$1.85$ percentage points and over LingBot-VA by $1.50$ points. The gain remains
consistent under domain randomization, where HiLo-WAM exceeds Fast-WAM by $1.72$
points and LingBot-VA by $2.02$ points.

On long-horizon tasks in RoboDojo, HiLo-WAM achieves an average score of $17.43$ and a success rate
of $6.50\%$. Its score improves over Fast-WAM by $8.29$ points and over
AHA-WAM by $8.82$ points, while the success rate increases by $1.33$ and
$3.83$ percentage points, demonstrating
a clear advantage on challenging long-horizon manipulation tasks.

\begin{table*}[t]
\centering

% ==================== RoboTwin 2.0 ====================
\begin{minipage}[t]{0.52\textwidth}
\vspace{0pt}
\centering
{\small\bfseries (a) RoboTwin~2.0\par}
\vspace{5pt}

\small
\setlength{\tabcolsep}{1pt}
\renewcommand{\arraystretch}{1.0}

\resizebox{\linewidth}{!}{%
\begin{tabular}{@{}lcccc@{}}
\toprule
\textbf{Model} & \textbf{Embodied PT.}
& \textbf{Clean} & \textbf{Rand.} & \textbf{Avg.} \\
\midrule

\multicolumn{5}{l}{\textit{VLA-based methods}} \\
$\pi_0$~\citep{black2026pi0visionlanguageactionflowmodel}
    & $\checkmark$ & 65.90 & 58.40 & 62.15 \\
X-VLA~\citep{zheng2026x}
    & $\checkmark$ & 72.90 & 72.80 & 72.85 \\
$\pi_{0.5}$~\citep{pmlr-v305-black25a}
    & $\checkmark$ & 82.70 & 76.80 & 79.75 \\
StarVLA~\citep{community2026starvlalegolikecodebasevisionlanguageaction}
    & $\times$ & 88.20 & 88.30 & 88.25 \\
ABot-M0~\citep{yang2026abot}
    & $\checkmark$ & 81.20 & 80.40 & 80.80 \\
LingBot-VLA~\citep{wu2026pragmatic}
    & $\checkmark$ & 86.50 & 85.30 & 85.90 \\

\midrule
\multicolumn{5}{l}{\textit{World-model-based methods}} \\
JEPA-VLA~\citep{miao2026jepa}
    & $\times$ & 73.50 & -- & -- \\
Motus~\citep{bi2026motus}
    & $\checkmark$ & 88.70 & 87.00 & 87.85 \\
LingBot-VA~\citep{li2026causal}
    & $\checkmark$ & 92.90 & 91.50
    & 92.20 \\
Fast-WAM~\citep{yuan2026fast}
    & $\times$ & 91.90 & 91.80 & 91.85 \\
Being-H0.7~\citep{luo2026being}
    & $\checkmark$ & 90.20 & 89.60 & 89.90 \\
AHA-WAM~\citep{cai2026aha}
    & $\times$ & \underline{93.40} & \underline{92.20} & \underline{92.80} \\
\midrule
\textbf{HiLo-WAM}
    & $\times$ & \textbf{93.87} & \textbf{93.52}
    & \textbf{93.70} \\
\bottomrule
\end{tabular}%
}
\end{minipage}%
\hfill
% ==================== RoboDojo ====================
\begin{minipage}[t]{0.46\textwidth}
\vspace{0pt}
\centering
{\small\bfseries (b) RoboDojo Long-Horizon\par}
\vspace{5pt}

\small
\setlength{\tabcolsep}{5pt}
\renewcommand{\arraystretch}{1.05}

\resizebox{\linewidth}{!}{%
\begin{tabular}{@{}lcc@{}}
\toprule
\textbf{Model} & \textbf{Score} & \textbf{Success (\%)} \\
\midrule

\multicolumn{3}{l}{\textit{VLA-based methods}} \\
OpenVLA-OFT~\citep{kim2025fine}
    & 0.70 & 0.00 \\
RDT~\citep{liu2025rdt}
    & 1.13 & 0.00 \\
LDA-1B~\citep{lyu2026lda}
    & 1.98 & 0.08 \\
H-RDT~\citep{bi2025hrdthumanmanipulationenhanced}
    & 2.23 & 0.17 \\
Pi-0~\citep{black2026pi0visionlanguageactionflowmodel}
    & 6.19 & 2.00 \\
GROOT-N1.7~\citep{bjorck2025gr00t}
    & 8.30 & 3.58 \\
LingBot-VLA~\citep{wu2026pragmatic}
    & 10.89 & 5.25 \\
StarVLA~\citep{community2026starvlalegolikecodebasevisionlanguageaction}
    & 14.15 & \textbf{6.50} \\
StarVLA-OFT
    & \underline{14.28} & \underline{6.17} \\

\midrule
\multicolumn{3}{l}{\textit{World-model-based methods}} \\
AHA-WAM~\citep{cai2026aha}
    & 8.61 & 2.67 \\
Fast-WAM~\citep{yuan2026fast}
    & 9.14 & 5.17 \\

\midrule
\textbf{HiLo-WAM}
    & \textbf{17.43} & \textbf{6.50} \\
\bottomrule
\end{tabular}%
}
\end{minipage}
\vspace{-5pt}
\caption{
\textbf{Evaluation on RoboTwin~2.0 and RoboDojo.}
(a) Average success rate (\%) across 50 RoboTwin~2.0 tasks
under clean and randomized conditions.
(b) Task progress score and average success rate (\%)
across 8 long-horizon RoboDojo tasks.
\emph{Embodied PT.} denotes embodied robot/action pretraining
before RoboTwin training.
For each metric, the best and second-best values among
the listed methods are shown in bold and underlined, respectively.
}
\label{tab:main_results}
\vspace{-10pt}
\end{table*}

\begin{figure*}[t]
    \centering
    \includegraphics[width=\textwidth]{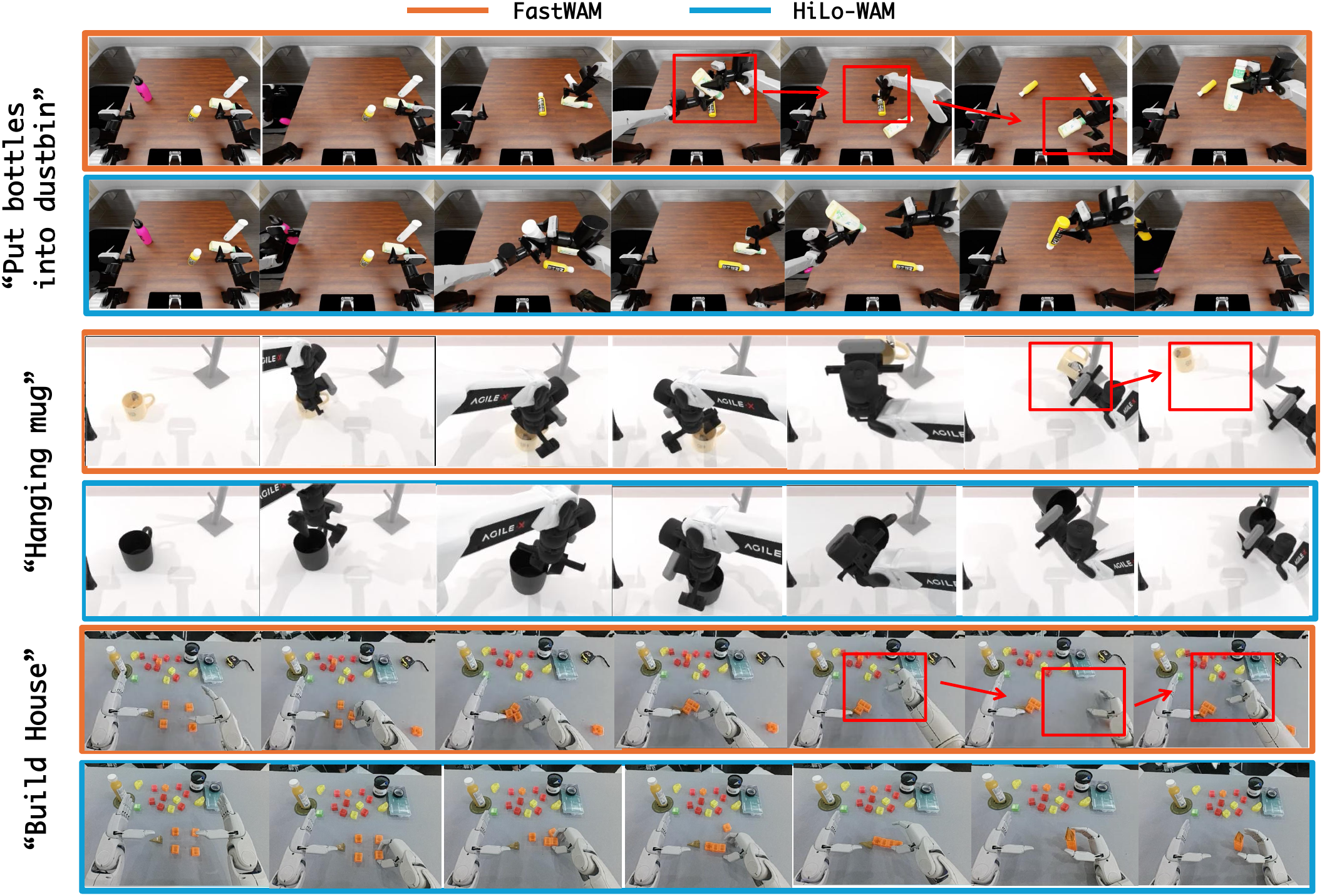}
    % \vspace{-10pt}
\caption{
\textbf{Qualitative comparison between Fast-WAM and HiLo-WAM in simulation and real-world tasks.}
Fast-WAM repeatedly changes its target bottle in \textit{Put Bottles into Dustbin}
and its target block in \textit{Build House}, while HiLo-WAM maintains coherent
manipulation targets throughout execution. In \textit{Mug Hanging}, Fast-WAM
fails because the mug handle is not properly aligned with the rack, whereas
HiLo-WAM successfully completes the final hanging operation.
}
    \label{fig:qualitative_results}
    \vspace{-15pt}
\end{figure*}

\paragraph{Qualitative Analysis.}
Figure~\ref{fig:qualitative_results} further illustrates HiLo-WAM's ability to capture both long-term task evolution and local dynamics in long-horizon manipulation. Specifically, in \textit{Put Bottles into Dustbin} and \textit{Build House}, Fast-WAM exhibits inconsistent task progression when faced with multimodal action distributions, whereas HiLo-WAM maintains coherent task execution. In \textit{Mug Hanging}, Fast-WAM fails to achieve the precise alignment required to complete the task, while HiLo-WAM successfully performs this fine-grained local interaction.

\subsection{Mechanism Analysis (Q2)}
To investigate how the decoupled visual dynamics contribute to action generation, we first quantify their respective contributions through ablation studies, then investigate the dependencies among the visual and action streams using attention maps and video visualizations.

\paragraph{Ablation Studies}
We first investigate how the visual dynamics at different temporal scales contribute to action generation.
Table~\ref{tab:robotwin_stream_ablation} ablates the historical and future
components of the long-horizon visual stream together with the local visual
dynamics stream on RoboTwin~2.0.

\begin{wraptable}{r}{0.55\textwidth}
% \vspace{-10pt}
\centering
\scriptsize
\setlength{\tabcolsep}{4pt}
\renewcommand{\arraystretch}{1.10}

\begin{tabular}{cccc}
\toprule
\multicolumn{2}{c}{\textbf{Long-Horizon stream}}
& \multirow{2}{*}{\shortstack{\textbf{Local Dynamics}\\\textbf{stream}}}
& \multirow{2}{*}{\textbf{Avg.\ SR}} \\
\cmidrule(lr){1-2}
\textbf{historical context} & \textbf{future evolution} & & \\
\midrule

\xmark & \xmark & \cmark & 90.58 \\
\xmark & \cmark & \cmark & 93.22 \\
\cmark & \cmark & \xmark & 93.52 \\
\midrule
\cmark & \cmark & \cmark & \textbf{93.70} \\

\bottomrule
\end{tabular}
% \vspace{-5pt}
\caption{
\textbf{RoboTwin~2.0 ablations.}
}
\label{tab:robotwin_stream_ablation}
\vspace{-10pt}
\end{wraptable}

Starting from the naive WAM with only the local dynamics stream, we introduce
the long-horizon stream while masking attention from the local dynamics
and action streams to $Z_t^{\mathrm{H,ctx}}$, allowing them to access
only $Z_t^{\mathrm{H,near}}$ within the long-horizon stream.
In this setting, historical context interacts bidirectionally only
within the long-horizon stream, providing memory for sparse future
prediction. This improves the average success rate from $90.58\%$
to $93.22\%$. Further enabling direct attention to historical context
increases it to $93.70\%$. Removing the local dynamics stream from
the full model reduces the success rate to $93.52\%$.
These results suggest that both long-horizon task progression
from sparse historical context and future evolution, and fine-grained
local dynamics, contribute to action generation.

\begin{figure*}[h]
    \centering
\vspace{-10pt}
    % Left figure
    \raisebox{0mm}{%
        \begin{minipage}[t]{0.49\textwidth}
            \vspace{0pt}
            \centering
            \includegraphics[width=\linewidth]{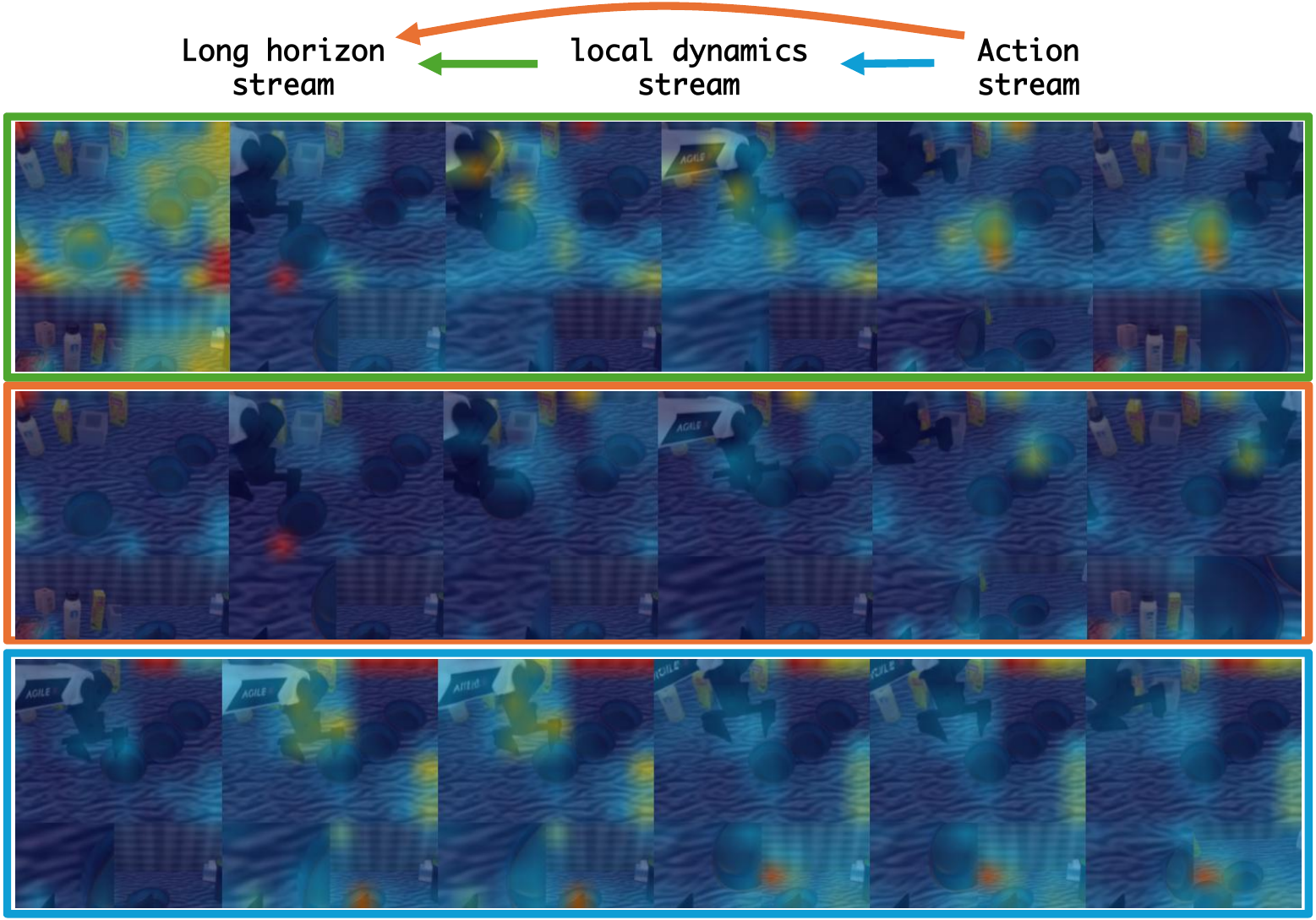}
            \vspace{-17pt}
            \captionof{figure}{
\textbf{Cross-stream attention map.}
Colored arrows indicate unidirectional attention maps between streams during a single inference process.
}
            \label{fig:attention_analysis}
        \end{minipage}
    }
    \hfill
    % Right figure
    \raisebox{-5mm}{%
        \begin{minipage}[t]{0.49\textwidth}
            \vspace{0pt}
            \centering
            \includegraphics[width=\linewidth]{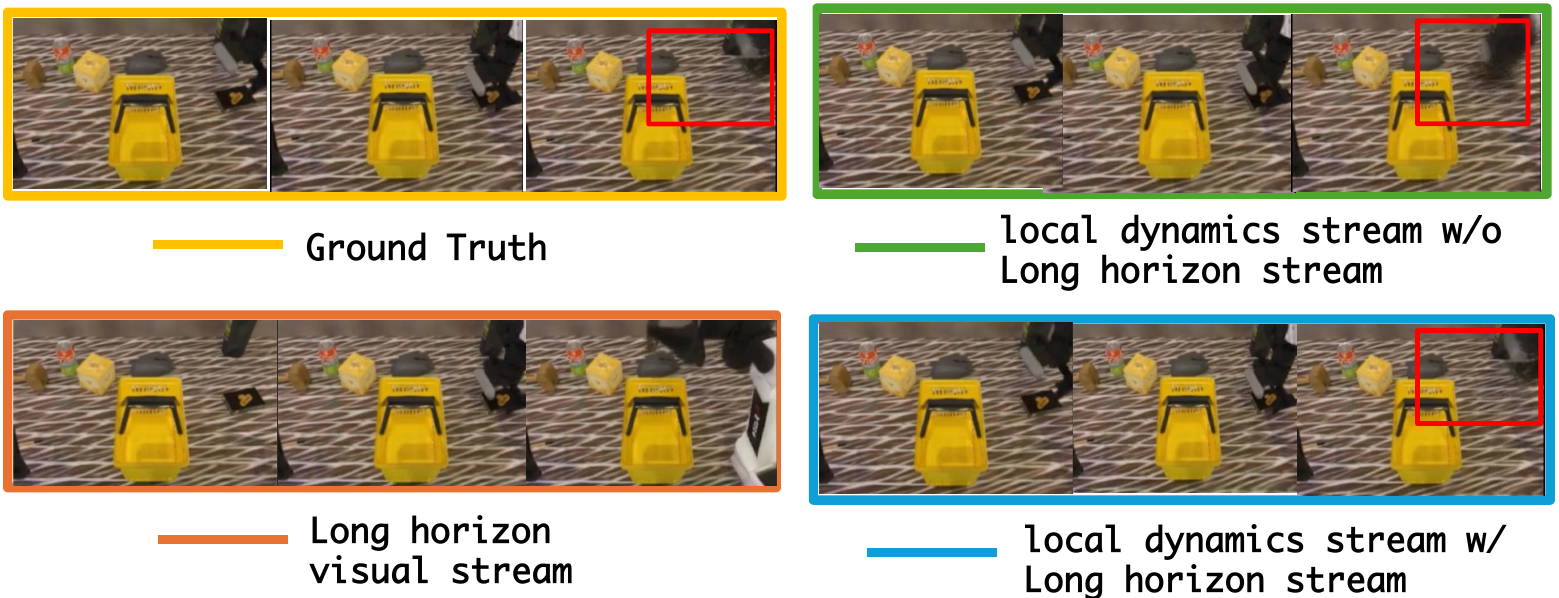}
            \captionof{figure}{
\textbf{Videos generated on \textit{Place Can Basket}.}
We show predictions from HiLo-WAM's long-horizon and local dynamics streams,
local dynamics predictions without the long-horizon stream, and the
corresponding local ground truth. Red boxes highlight the key
interaction regions.
}
            \label{fig:local_guidance}
        \end{minipage}
    }
\vspace{-10pt}
\end{figure*}

\paragraph{Dependency Analysis}
% \paragraph{Cross-stream interaction analysis.}
We further investigate the underlying mechanisms by analyzing dependencies among the different streams.
We first visualize the attention maps between different streams for the \textit{Stack Bowls Three} task in Figure~\ref{fig:attention_analysis}. We observe that the long-horizon stream enables both the local dynamics and action streams to perceive changes in robot-arm and object positions over an extended horizon. The local dynamics stream further captures fine-grained robot-arm motion and dynamic robot--object interactions for the action stream. Together, the two visual streams guide action generation to account for both long-horizon task progression and local dynamics.

We further illustrate the effect of this dependencies through the generated videos in Figure~\ref{fig:local_guidance}. In the \textit{Place Can Basket} task, the long-horizon stream predicts a sparse grasp-to-place sequence from sparse historical context, guiding the local dynamics stream to generate a short-horizon grasp video aligned with the overall task progression. Without this guidance, local predictions become less task-directed, with no clear arm motion toward the intended manipulation. Overall, HiLo-WAM explicitly models the dependencies among task evolution, local dynamics, and action generation, demonstrating their mutually reinforcing effects.

\subsection{Effect of Temporal Scale (Q3)}

\begin{wrapfigure}{r}{0.4\textwidth}
    \centering
    \vspace{-30pt}
    \includegraphics[width=\linewidth]{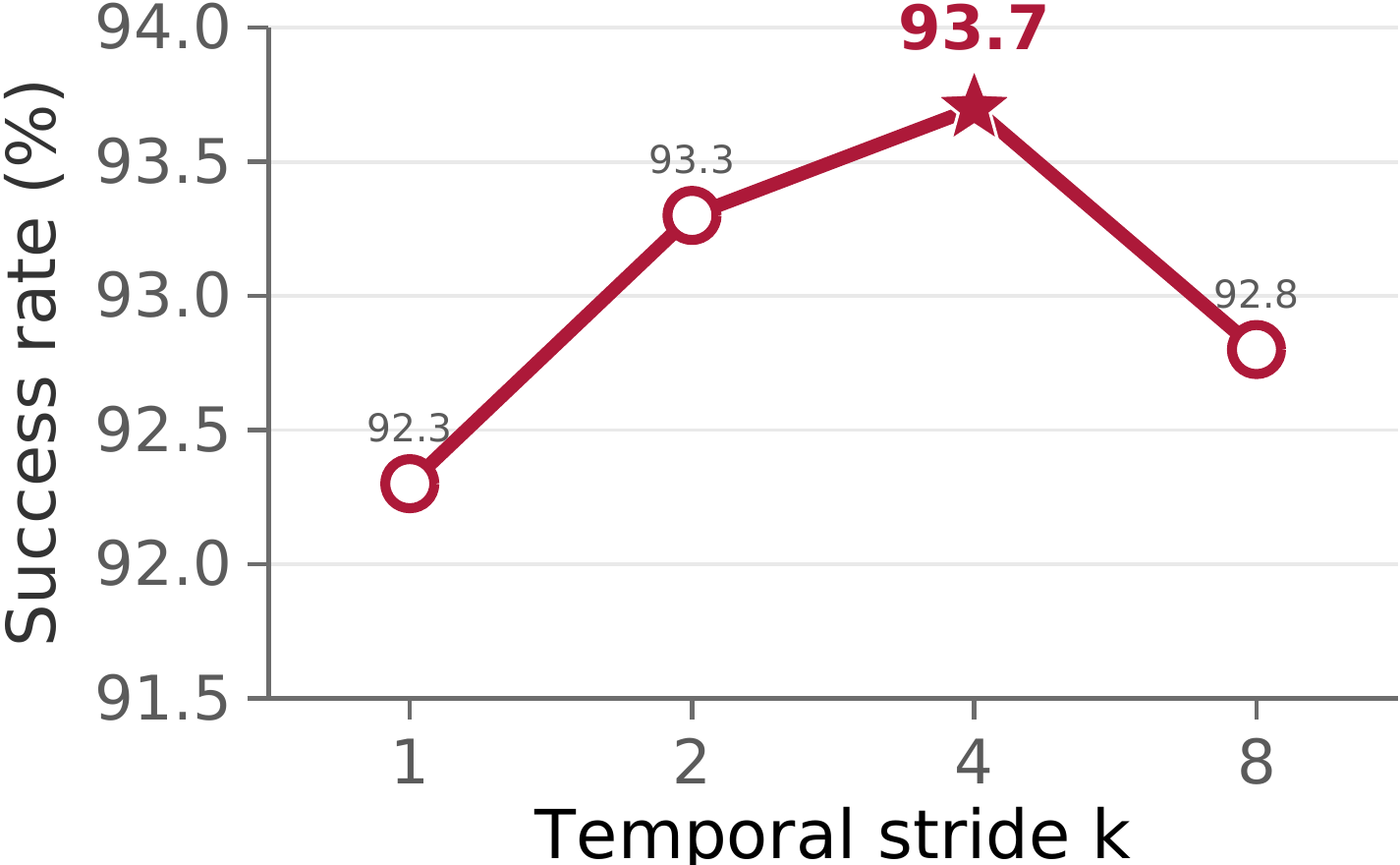}
    \vspace{-15pt}
    \caption{\textbf{RoboTwin~2.0 success rate with different temporal strides.}}
    \label{fig:scale_ablation}
    \vspace{-20pt}
\end{wrapfigure}
We analyze how different temporal scales of the long-horizon visual stream affect manipulation performance. With a fixed number of visual tokens, we vary the temporal stride $k$ to construct long-horizon dynamics with different temporal coverage and temporal resolution.

As shown in Fig.~\ref{fig:scale_ablation}, performance does not increase
monotonically with temporal stride. Increasing $k$ from $1$ to $4$ improves
the success rate from $92.3\%$ to $93.7\%$, suggesting that broader temporal
coverage helps capture long-horizon task evolution. However, further increasing
$k$ to $8$ reduces performance to $92.8\%$, as the visual observations become
too sparse to preserve sufficient temporal information. These results indicate
that long-horizon world modeling benefits from an appropriate temporal scale
that balances temporal coverage and temporal resolution, with $k=4$ achieving
the best trade-off in our experiments.

\subsection{Inference efficiency (Q4)}
\label{sec:infer_efficiency}

To further evaluate the efficiency of progressive inference, we report the end-to-end latency
of a single action-chunk inference call and success rate on RoboTwin~2.0 using a single NVIDIA RTX PRO 5000 GPU. The prompt is encoded once at startup, and its embeddings are cached and reused during subsequent steady-state inference,
and the measured latency mainly includes Video VAE processing, Video DiT
denoising, and Action DiT denoising.  
Table~\ref{tab:robotwin_latency} reports the inference latency and performance of each configuration. The full denoising schedule consists
of $10$ steps for both video and action streams. Across all configurations,
the action stream retains the full schedule with $n_{\mathrm A}=10$.
When joint denoising is disabled, all the streams are denoised sequentially, with each completed
stream's features cached for subsequent streams.

\begin{table}[h]
\centering
\small
\setlength{\tabcolsep}{4pt}
\renewcommand{\arraystretch}{1.05}
\begin{tabular}{lcccrrc}
\toprule
\textbf{Configuration}
& \textbf{$n_{\mathrm H}/n_{\mathrm L}$}
& \textbf{Joint}
& \textbf{Skipping}
& \textbf{Lat. (ms)}
& \textbf{Speedup}
& \textbf{SR (\%)} \\
\midrule
Baseline
& 10 / 10 &  &  & 1114.46 & 1.00$\times$ & 93.54 \\
+ Partial denoising
& 5 / 5 &  &  & 723.65 & 1.54$\times$ & 93.62 \\
+ Joint denoising
& 5 / 5 & $\checkmark$ &  & 619.85 & 1.80$\times$ & 93.70 \\
+ Dynamic skipping
& 5 / 5 & $\checkmark$ & $\checkmark$
& \textbf{380.02} & \textbf{2.93$\times$} & 93.70 \\
\bottomrule
\end{tabular}
\vspace{-5pt}
\caption{
\textbf{Inference Efficiency and Performance.}
Speedup is measured relative to the baseline.
}
\label{tab:robotwin_latency}
\vspace{-5pt}
\end{table}

\begin{figure*}[h]
    \centering
    % \vspace{-15pt}
    \includegraphics[width=\textwidth]{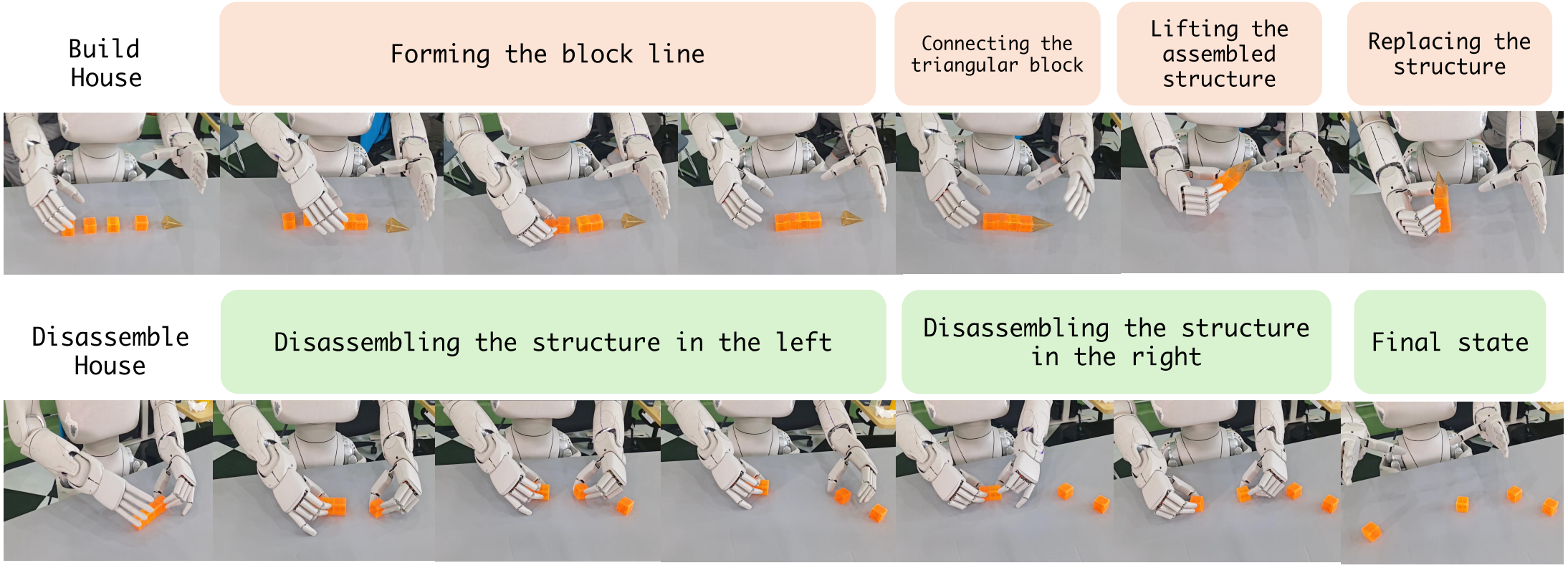}
    \vspace{-15pt}
    \caption{\textbf{Detailed task progressions of real-world tasks on the Fourier GR-3.}
    Each task involves a sequence of operations on multiple blocks, including assembly, disassembly, lifting, and placement, to meet specific shape requirements.}
    \label{fig:realworld_tasks}
\end{figure*}

As shown in Table~\ref{tab:robotwin_latency}, partial denoising reduces
latency from $1114.46$ to $723.65$\,ms while improving success rate
from $93.54\%$ to $93.62\%$. Joint denoising further lowers latency
to $619.85$\,ms and increases performance to $93.70\%$.
Dynamic skipping further reduces latency to $380.02$\,ms.
Overall, these strategies achieve a $2.93\times$ speedup over the baseline while improving performance.

\subsection{Real-World Robot Experiments}
\label{sec:real_results}

\paragraph{Experiment Setup.}
We compare HiLo-WAM with $\pi_{0.5}$~\citep{pmlr-v305-black25a} and
Fast-WAM~\citep{yuan2026fast} on three long-horizon real-world tasks, with 10 trials per task.
OOD evaluation includes changes in lighting, tabletop background, and clutter.
As shown in Figure~\ref{fig:realworld_tasks}, a rollout is considered successful if the task is completed within the fixed
episode horizon. To measure partial progress, each task is divided into ten
ordered milestones, yielding a progress score from 0 to 10. We report the mean
success rate and progress score over all trials, with task definitions and
milestone criteria provided in the appendix~\ref{app:realworld}.
\begin{figure*}[h]
    \centering
    \vspace{-7pt}
    \includegraphics[width=\textwidth]{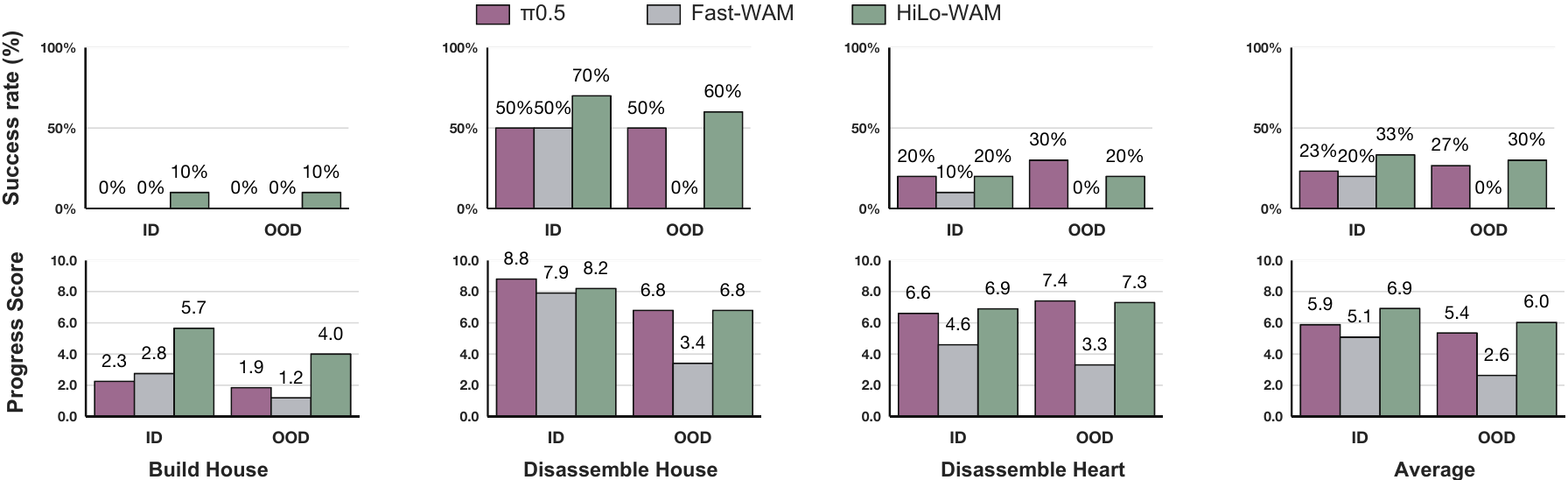}
    \vspace{-15pt}
\caption{
\textbf{Real-world evaluation on the Fourier GR-3 robot.}
}
    \label{fig:realworld_results}
    \vspace{-7pt}
\end{figure*}

\paragraph{Results.}
As shown in Figure~\ref{fig:realworld_results}, HiLo-WAM achieves the strongest performance in both in-domain and OOD settings. In-domain, it achieves an average score of $6.9$ and a success rate of $33\%$, surpassing $\pi_{0.5}$ by $1.0$ score points and $10$ percentage points, and Fast-WAM by $1.8$ score points and $13$ percentage points. Under OOD conditions, it reaches $6.0$ and $30\%$, exceeding $\pi_{0.5}$ by $0.6$ score points and $3$ percentage points, and Fast-WAM by $3.4$ score points and $30$ percentage points. This validates the stability and robustness of HiLo-WAM on real-world long-horizon manipulation tasks.

\section{Conclusion}

We introduced \textbf{HiLo-WAM}, a World Action Model that separates
long-horizon task evolution from fine-grained local interaction dynamics.
HiLo-WAM models these two complementary processes at different temporal scales
and coordinates them through structured cross-stream interaction for action
generation. Progressive stream-wise denoising further limits the additional
inference cost introduced by long-horizon visual modeling.
Experiments in simulation and on a physical robot demonstrate strong
long-horizon manipulation performance. Ablations show that long-horizon and
local visual information contribute differently to action generation, while
temporal-scale analysis highlights the importance of balancing temporal
coverage and resolution. These results support decoupled multi-scale visual
world modeling as an effective direction for long-horizon robotic manipulation.

\paragraph{Limitations.}
HiLo-WAM improves long-horizon manipulation by modeling visual dynamics at
different temporal scales within the World Action Model. However, it remains
limited by the semantic understanding and predictive capability of the
underlying video model. Complex long-horizon tasks that require explicit
subtask decomposition, progress estimation, or higher-level reasoning may
therefore remain challenging. Future work could incorporate stronger semantic
models, such as VLMs, to enhance task understanding, reasoning, and
generalization.

\section*{AI Use Statement}

In this work, we did not use generative AI tools for research methodology or
experimental design, theoretical development, mathematical derivations, or
other scientific reasoning; these aspects were independently completed by the
authors. We used generative AI tools only for language polishing and
presentation refinement, including improving the clarity, fluency, and
readability of author-written text. All AI-assisted content was carefully
reviewed, verified, and revised by the authors. We take responsibility for the
final content of this work, including text, claims, or artifacts produced with
the aid of generative AI.

\section*{Reproducibility Statement}

We provide sufficient implementation and experimental details to facilitate
reproduction of our results. The model architecture, training objectives,
inference procedure, and key hyperparameters are described in the main paper
and appendix. Detailed evaluation protocols for RoboTwin~2.0, RoboDojo, and
the real-world experiments, including task definitions, OOD settings, and
milestone-based progress metrics, are also provided in the appendix.


% \newpage

\bibliography{iclr2027_conference}
\bibliographystyle{iclr2027_conference}

\newpage
\appendix
% \textcolor{blue}{Code can be found in the Supplementary Materials.}

% \textcolor{blue}{More videos can be found at the anonymous website: } \url{https://hilowam.github.io/}.

\section{Training and Implementation Details}
\label{app:training}

We follow Fast-WAM~\cite{yuan2026fast} for data processing and backbone configuration, and
summarize the additional training and inference configurations introduced
for multi-scale modeling in Table~\ref{tab:implementation_details}.

\begin{table*}[h]
\centering
\small
\setlength{\tabcolsep}{7pt}
\renewcommand{\arraystretch}{1.08}
\begin{tabular}{lll}
\toprule
\textbf{Category} & \textbf{Parameter} & \textbf{Value} \\
\midrule

\multirow{9}{*}{Model}
& Video backbone & Wan2.2-TI2V-5B \\
& Text tokenizer & Wan2.1-T2V-1.3B \\
& Video DiT layers & 30 \\
& Video DiT hidden / FFN dim. & 3072 / 14336 \\
& Action DiT layers & 30 \\
& Action DiT hidden / FFN dim. & 1024 / 4096 \\
& Video latent patch size & $1 \times 2 \times 2$ \\
\midrule

\multirow{7}{*}{Data \& Temporal Modeling}
& image resolution & $240 \times 320$ \\
& Action horizon $H$ & 32 \\
&Video chunk $T$ & 32 \\
& Action / video frequency ratio & 4 \\
\midrule

\multirow{4}{*}{Long-Horizon Stream}
& Historical keyframes $m$ & 8 \\
& Future keyframes $n$ & 8 \\
& Temporal stride $k$ & 4 \\
\midrule

\multirow{4}{*}{Robot Representation}
& Action dimension & 14 \\
& Proprioceptive state dimension & 14 \\
& Action/state normalization & Z-score \\
\midrule

\multirow{8}{*}{Training}
& Learning rate & $1\times10^{-4}$ \\
& LR scheduler & Cosine \\
& Weight decay & $1\times10^{-2}$ \\
& Gradient accumulation steps & 1 \\
& Action loss weight $\lambda_{\mathrm A}$ & 1.0 \\
& Long-horizon stream loss weight $\lambda_{\mathrm H}$ & 1.0 \\
& Local dynamics stream loss weight $\lambda_{\mathrm L}$ & 1.0 \\
\midrule

\multirow{3}{*}{Noise Scheduler}
& Video train / inference shift & 6.0 / 6.0 \\
& Action train / inference shift & 3.0 / 3.0 \\
& Separated timestep conditioning & Enabled \\
\midrule

\multirow{7}{*}{Inference}
& $n_H$ & 5\\
& $n_L$ & 5\\
& $n_A$ & 10\\
& Joint denoising & Enabled \\
& Text CFG scale & 1.0 \\
\bottomrule
\end{tabular}
\caption{
\textbf{Training and inference configurations of HiLo-WAM.}
We report the main architectural, data, optimization, and inference
hyperparameters used in our experiments.
}
\label{tab:implementation_details}
\end{table*}

\section{Simulation Benchmark Details}
\label{app:simulation}
We report the per-task success rates on RoboTwin~2.0 under different temporal
stride settings $k$, as shown in Table~\ref{tab:robotwin_per_task}. Here,
HiLo-WAM-$k2$, HiLo-WAM-$k8$ denote variants using
temporal strides of $k=2$, $8$, respectively, for the long-horizon
visual stream.

\begin{table}[t]
\centering
\scriptsize
\setlength{\tabcolsep}{2.8pt}
\renewcommand{\arraystretch}{0.92}
\resizebox{\textwidth}{!}{
\begin{tabular}{l cc cc cc cc cc cc}
\toprule
\textbf{Task}
& \multicolumn{2}{c}{\textbf{Fast-WAM}}
& \multicolumn{2}{c}{\textbf{LingBot-VA}}
& \multicolumn{2}{c}{\boldmath$\pi_{0.5}$}
& \multicolumn{2}{c}{\textbf{Motus}}
& \multicolumn{2}{c}{\textbf{HiLo-WAM-$k2$}}
& \multicolumn{2}{c}{\textbf{HiLo-WAM-$k8$}} \\
\cmidrule(lr){2-3}
\cmidrule(lr){4-5}
\cmidrule(lr){6-7}
\cmidrule(lr){8-9}
\cmidrule(lr){10-11}
\cmidrule(lr){12-13}
& Clean & Rand.
& Clean & Rand.
& Clean & Rand.
& Clean & Rand.
& Clean & Rand.
& Clean & Rand. \\
\midrule

Adjust Bottle
& \textbf{100} & \textbf{100}
& 90 & 94
& \textbf{100} & 99
& 89 & 93
& 98 & \textbf{100}
& 94 & 98 \\

Beat Block Hammer
& 99 & 97
& 96 & \textbf{98}
& 96 & 93
& 95 & 88
& \textbf{100} & 96
& 96 & 92 \\

Blocks Ranking RGB
& \textbf{100} & \textbf{100}
& 99 & 98
& 92 & 85
& 99 & 97
& \textbf{100} & \textbf{100}
& \textbf{100} & 92 \\

Blocks Ranking Size
& 94 & \textbf{98}
& 94 & 96
& 49 & 26
& 75 & 63
& \textbf{96} & 94
& 88 & 88 \\

Click Alarmclock
& \textbf{100} & \textbf{100}
& 99 & \textbf{100}
& 98 & 89
& \textbf{100} & \textbf{100}
& \textbf{100} & \textbf{100}
& \textbf{100} & \textbf{100} \\

Click Bell
& \textbf{100} & \textbf{100}
& \textbf{100} & \textbf{100}
& 99 & 66
& \textbf{100} & \textbf{100}
& \textbf{100} & \textbf{100}
& \textbf{100} & \textbf{100} \\

Dump Bin Bigbin
& 97 & 96
& 89 & 96
& 92 & 97
& 95 & 91
& 98 & \textbf{100}
& \textbf{100} & \textbf{100} \\

Grab Roller
& \textbf{100} & \textbf{100}
& \textbf{100} & \textbf{100}
& \textbf{100} & \textbf{100}
& \textbf{100} & \textbf{100}
& \textbf{100} & \textbf{100}
& \textbf{100} & \textbf{100} \\

Handover Block
& 95 & 81
& \textbf{99} & 78
& 66 & 57
& 86 & 73
& 92 & 86
& 96 & \textbf{92} \\

Handover Mic
& 99 & \textbf{100}
& 94 & 96
& 98 & 97
& 78 & 63
& 92 & \textbf{100}
& \textbf{100} & 98 \\

Hanging Mug
& 58 & \textbf{62}
& 40 & 28
& 18 & 17
& 38 & 38
& 46 & \textbf{62}
& \textbf{62} & \textbf{62} \\

Lift Pot
& \textbf{100} & \textbf{100}
& \textbf{100} & 99
& 96 & 85
& 96 & 99
& \textbf{100} & \textbf{100}
& \textbf{100} & \textbf{100} \\

Move Can Pot
& 90 & 88
& 94 & \textbf{97}
& 51 & 55
& 34 & 74
& \textbf{100} & 96
& 92 & 96 \\

Move Pillbottle Pad
& \textbf{100} & 99
& 99 & 99
& 84 & 61
& 93 & 96
& \textbf{100} & \textbf{100}
& \textbf{100} & \textbf{100} \\

Move Playingcard Away
& \textbf{100} & \textbf{100}
& \textbf{100} & 99
& 96 & 84
& \textbf{100} & 96
& \textbf{100} & \textbf{100}
& \textbf{100} & \textbf{100} \\

Move Stapler Pad
& 77 & 64
& \textbf{91} & 79
& 56 & 42
& 83 & \textbf{85}
& 72 & 78
& 72 & 52 \\

Open Laptop
& \textbf{98} & \textbf{100}
& 92 & 94
& 90 & 96
& 95 & 91
& \textbf{98} & 98
& \textbf{98} & \textbf{100} \\

Open Microwave
& 62 & 45
& 82 & 86
& 34 & 77
& 95 & \textbf{91}
& 86 & 90
& \textbf{100} & 90 \\

Pick Diverse Bottles
& 80 & 85
& 89 & 82
& 81 & 71
& \textbf{90} & \textbf{91}
& 84 & 84
& 84 & 86 \\

Pick Dual Bottles
& \textbf{100} & 96
& \textbf{100} & \textbf{99}
& 93 & 63
& 96 & 90
& 98 & 94
& \textbf{100} & 96 \\

Place A2B Left
& 95 & 93
& \textbf{97} & 93
& 87 & 82
& 88 & 79
& 94 & \textbf{94}
& 90 & \textbf{94} \\

Place A2B Right
& 93 & \textbf{99}
& \textbf{97} & 95
& 87 & 84
& 91 & 87
& 94 & 96
& 92 & 94 \\

Place Bread Basket
& 91 & 93
& 97 & \textbf{95}
& 77 & 64
& 91 & 94
& 96 & 94
& \textbf{98} & 94 \\

Place Bread Skillet
& 90 & \textbf{93}
& \textbf{95} & 90
& 85 & 66
& 86 & 83
& 90 & 90
& 92 & 92 \\

Place Burger Fries
& 96 & 99
& 97 & 95
& 94 & 87
& 98 & 98
& \textbf{100} & \textbf{100}
& 96 & 98 \\

Place Can Basket
& 71 & 69
& 81 & \textbf{84}
& 62 & 62
& 81 & 76
& \textbf{84} & 82
& 72 & 78 \\

Place Cans Plasticbox
& 99 & 96
& \textbf{100} & 99
& 94 & 84
& 98 & 94
& \textbf{100} & \textbf{100}
& 98 & 96 \\

Place Container Plate
& 96 & \textbf{100}
& \textbf{99} & 97
& \textbf{99} & 95
& 98 & 99
& 98 & 98
& 98 & \textbf{100} \\

Place Dual Shoes
& 94 & 88
& 94 & 89
& 75 & 75
& 93 & 87
& \textbf{96} & \textbf{92}
& 88 & 90 \\

Place Empty Cup
& \textbf{100} & \textbf{100}
& \textbf{100} & \textbf{100}
& \textbf{100} & 99
& 99 & 98
& \textbf{100} & \textbf{100}
& \textbf{100} & \textbf{100} \\

Place Fan
& 96 & 96
& \textbf{99} & 93
& 87 & 85
& 91 & 87
& 98 & 96
& 92 & \textbf{100} \\

Place Mouse Pad
& 83 & 89
& 93 & 96
& 60 & 39
& 66 & 68
& \textbf{100} & \textbf{98}
& 98 & \textbf{98} \\

Place Object Basket
& 89 & 88
& \textbf{91} & 88
& 80 & 76
& 81 & 87
& 88 & \textbf{90}
& 78 & 78 \\

Place Object Scale
& 90 & 97
& \textbf{96} & 95
& 86 & 80
& 88 & 85
& \textbf{96} & 94
& \textbf{96} & \textbf{98} \\

Place Object Stand
& 90 & 94
& \textbf{99} & 96
& 91 & 85
& 98 & \textbf{97}
& 90 & 88
& 88 & 94 \\

Place Phone Stand
& 97 & 99
& 97 & 97
& 81 & 81
& 87 & 86
& 94 & 98
& \textbf{98} & \textbf{100} \\

Place Shoe
& 96 & \textbf{99}
& 98 & 98
& 92 & 93
& \textbf{99} & 97
& 94 & 96
& 98 & 92 \\

Press Stapler
& 90 & 97
& 85 & 82
& 87 & 83
& 93 & \textbf{98}
& \textbf{96} & \textbf{98}
& \textbf{96} & 94 \\

Put Bottles Dustbin
& \textbf{95} & 90
& 87 & 91
& 84 & 79
& 81 & 79
& 92 & \textbf{98}
& 82 & 92 \\

Put Object Cabinet
& \textbf{94} & 89
& 85 & 87
& 80 & 79
& 88 & 71
& 90 & 88
& 90 & \textbf{90} \\

Rotate QRCode
& 93 & 89
& \textbf{96} & \textbf{91}
& 89 & 87
& 89 & 73
& 90 & 82
& 86 & 84 \\

Scan Object
& 89 & \textbf{92}
& \textbf{96} & 91
& 72 & 65
& 67 & 66
& 80 & \textbf{92}
& 94 & \textbf{92} \\

Shake Bottle
& \textbf{100} & \textbf{100}
& \textbf{100} & 97
& 99 & 97
& \textbf{100} & 97
& \textbf{100} & \textbf{100}
& \textbf{100} & \textbf{100} \\

Shake Bottle Horizontally
& \textbf{100} & \textbf{100}
& \textbf{100} & 99
& 99 & 99
& \textbf{100} & 98
& \textbf{100} & \textbf{100}
& \textbf{100} & \textbf{100} \\

Stack Blocks Three
& 95 & 97
& 99 & 98
& 91 & 76
& 91 & 95
& \textbf{100} & 96
& 98 & \textbf{100} \\

Stack Blocks Two
& \textbf{100} & \textbf{100}
& \textbf{100} & 98
& 97 & \textbf{100}
& \textbf{100} & 98
& \textbf{100} & \textbf{100}
& \textbf{100} & 98 \\

Stack Bowls Three
& 80 & 81
& 86 & 83
& 77 & 71
& 79 & 87
& 68 & 80
& \textbf{90} & \textbf{88} \\

Stack Bowls Two
& 92 & \textbf{98}
& 94 & \textbf{98}
& 95 & 96
& \textbf{98} & \textbf{98}
& 92 & 94
& 88 & 96 \\

Stamp Seal
& 90 & 94
& 96 & 97
& 79 & 55
& 93 & 92
& \textbf{100} & 96
& 96 & \textbf{100} \\

Turn Switch
& 61 & 59
& 44 & 45
& 62 & 54
& \textbf{84} & \textbf{78}
& 74 & 72
& 64 & 66 \\

\midrule
\textbf{Average}
& 91.88 & 91.78
& 92.90 & 91.50
& 82.74 & 76.76
& 88.66 & 87.02
& \textbf{93.08} & \textbf{93.60}
& 92.76 & 92.76 \\
\bottomrule
\end{tabular}
}
\caption{
\textbf{Per-task success rates (\%) on RoboTwin~2.0.}
We report results under both clean and randomized settings.
The best result for each task and setting is highlighted in bold.
}
\label{tab:robotwin_per_task}
\end{table}

\section{Real-World Experiment Details}
\label{app:realworld}

% \subsection{Real-World Task Definitions and Progress Scoring}
% \label{app:realworld_tasks}

We evaluate three long-horizon manipulation tasks on the physical robot:
\textit{Build House}, \textit{Disassemble House}, and
\textit{Disassemble Heart}. Each task consists of multiple sequential
manipulation stages and is assigned a progress score ranging from 0 to 10.
The score is accumulated according to the completion of predefined
intermediate stages, allowing us to measure partial task progress in addition
to binary task success. A score of 10 corresponds to full task completion.
The detailed scoring criteria are described below.

\paragraph{Build House.}
The robot is instructed to arrange four orange blocks into a straight line,
connect a brown triangular block to the assembled line, lift the complete
structure, and finally place it back on the table. The total score is divided
into four stages:
\begin{itemize}
    \item \textbf{Forming the block line (4 points).}
    Up to 4 points are awarded according to the completion and quality of the
    straight-line arrangement of the four orange blocks. Correctly connected
    and aligned blocks receive full credit, while connected but visibly bent
    arrangements receive partial credit.

    \item \textbf{Connecting the triangular block (2 points).}
    One point is awarded when the triangular block is moved to the designated
    end of the line with the correct orientation, and an additional point is
    awarded once a stable connection is established.

    \item \textbf{Lifting the assembled structure (2 points).}
    One point is awarded for establishing a controlled grasp on the assembled
    structure. The remaining point is awarded when the complete structure is
    lifted from the table and remains intact during the lift.

    \item \textbf{Replacing the structure (2 points).}
    One point is awarded when the robot places the structure back on the table
    and releases it in a controlled manner. An additional point is awarded if
    the structure remains connected and stable for at least two seconds after
    release.
\end{itemize}

\paragraph{Disassemble House.}
The robot first grasps the orange block, transfers it between the two hands,
disassembles the connected structure into individual components, and places
the separated pieces into the designated left and right regions of the table.
The scoring criteria are:
\begin{itemize}
    \item \textbf{Disassembling the structure (6 points).}
    The score is assigned proportionally to the number of required connections
    that are successfully separated. Full separation of all components yields
    6 points. For example, when three connections must be broken, each
    successfully separated connection contributes 2 points.

    \item \textbf{Placement into left and right regions (2 points).}
    One point is awarded when both sides contain at least one separated block,
    and the second point is awarded when all components are placed inside
    their predefined target regions.

    \item \textbf{Final state (2 points).}
    Full credit is awarded when all components remain mutually separated and
    stationary for at least two seconds, with no block leaving the designated
    workspace.
\end{itemize}

\paragraph{Disassemble Heart.}
The robot is required to first remove the two yellow blocks from the red
interlocking structure and place them on the table, and then disassemble the
remaining red structure into individual components. The scoring criteria are:
\begin{itemize}
    \item \textbf{Removing the yellow blocks (2 points).}
    Each yellow block contributes 1 point when it is successfully removed and
    stably placed on the table. A block that is removed but unintentionally
    dropped receives partial credit of 0.5 points.

    \item \textbf{Disassembling the red structure (6 points).}
    The score is proportional to the number of required connections that are
    successfully separated. Completely separating all red blocks yields
    6 points. For example, when three connections are present, each separated
    connection contributes 2 points.

    \item \textbf{Final state (2 points).}
    Full credit is awarded when the robot has released the objects and all
    yellow and red blocks remain stable for at least two seconds without any
    object leaving the workspace.
\end{itemize}

\end{document}